# A Case Study on Evaluating Genetic Algorithms for Early Building Design Optimization: Comparison with Random and Grid Searches

Farnaz Nazari[1], Wei Yan[2]

[1] Corresponding author: PhD, Department of Construction Science, Texas A&M University; email: farnaz.nazari@tamu.edu.
[2] Professor, Department of Architecture, Texas A&M University.

## Abstract

In early-stage architectural design, optimization algorithms are essential for efficiently exploring large and complex design spaces under tight computational constraints. While prior research has benchmarked various optimization methods, their findings often lack generalizability to real-world, domain-specific problems—particularly in early building design optimization for energy performance. This study evaluates the effectiveness of Genetic Algorithms (GAs) for early design optimization, focusing on their ability to find near-optimal solutions within limited timeframes. Using a constrained case study, we compare a simple GA to two baseline methods—Random Search (RS) and Grid Search (GS) —with each algorithm tested 10 times to enhance the reliability of the conclusions. Our findings show that while RS may miss optimal solutions due to its stochastic nature, it was unexpectedly effective under tight computational limits. Despite being more systematic, GS was outperformed by RS, likely due to the irregular design search space. This suggests that, under strict computational constraints, lightweight methods like RS can sometimes outperform more complex approaches like GA. As this study is limited to a single case under specific constraints, future research should investigate a broader range of design scenarios and computational settings to validate and generalize the findings. Additionally, the potential of Random Search or hybrid optimization methods should be further investigated, particularly in contexts with strict computational limitations.

# 1   Introduction

Optimization is fundamental in building and architectural design research, where balancing competing performance measures is often necessary [1], [2]. While substantial research exists for benchmarking different optimization methods, their conclusions may not universally apply to all real-world problems [3]. Under certain constraints, the evaluation and benchmarking of optimization methods can be misleading, potentially masking an algorithm's strengths or weaknesses, or suggesting inappropriate algorithmic choices for specific situations [3], as no single algorithm consistently outperforms others across all performance metrics in building energy simulation problems [4], [5]. Additionally, understanding the fitness landscape of each optimization problem is crucial for developing algorithms tailored to its unique challenges [6]. Therefore, it is essential to select the appropriate method based on the problem's specific needs and context.

In the early stages of architectural design, there is a critical need for search algorithms capable of efficiently navigating extensive solution spaces for a trade-off between accuracy and speed [7], [8], [9]. These algorithms must identify solutions that are close to optimal within a constrained timeframe, enabling the design team to explore and conceptualize design ideas effectively and promptly [10], [11]. For example, factors like building shape, including self-shading forms, influence energy performance [12], [13], yet their impact is often underemphasized early on, leading later stages to focus more on refining the initial shape rather than optimizing it holistically [14], [15], [16].

Genetic algorithm (GA) and their variants have proven to be effective tools for solving complex optimization problems in architectural design [17], [10], [18], [19], [20]. Their capability to efficiently explore large design spaces positions them as viable options for achieving high-performance solutions, especially when a trade-off between accuracy and computational effort is acceptable, particularly advantageous in scenarios with limited evaluation budgets [21], [22], [23], [24]. However, when computational resources are excessively constrained, even GA may struggle to deliver effective solutions, particularly due to the risk of premature convergence and getting trapped in local optima [4] [25] [6] [6], [26]. While GAs excel at exploring large solution spaces, their performance is highly dependent on computational resources and proper parameter tuning [4] [25], along with fitness landscape and optimization variables [6] [26]. The risk of GA getting trapped in local optima increases depending on the fitness landscape and the variables being optimized, especially when the number of simulations is limited [6].

This paper evaluates the effectiveness of GA in optimizing building design, focusing on their strengths and limitations in early-stage architectural design. Through a case study with computational constraints, it compares the performance of a simple GA against baseline methods like Random and Grid Searches to assess GA's viability under strict computational limitations. The ultimate goal is to provide insights that inform the development of more efficient, hybrid optimization strategies tailored to the unique challenges of building design.

The paper is structured as follows: Section 2 reviews the background literature; Section 3 defines the test case; Section 4 explores the problem's fitness landscape; Section 5 compares the performance of three optimization methods; and Section 6 presents the analysis and discussion of the results.

## 2 Background

### 2.1 Genetic Algorithm application in Building Design Optimization

Genetic Algorithms (GAs) are a widely used class of stochastic optimization algorithms in building applications [27]. Review studies, including Nguyen et al.'s analysis of over 200 papers in building optimization, GAs were found to be used in 40% of cases, followed by particle swarm optimization (13%), hybrid algorithms (10%). [17], [28]. The basic idea behind GA is to simulate natural selection by evolving a population of solutions through crossover and mutation, enabling both local and global search [29]. Figure 1 illustrates its iterative optimization process.

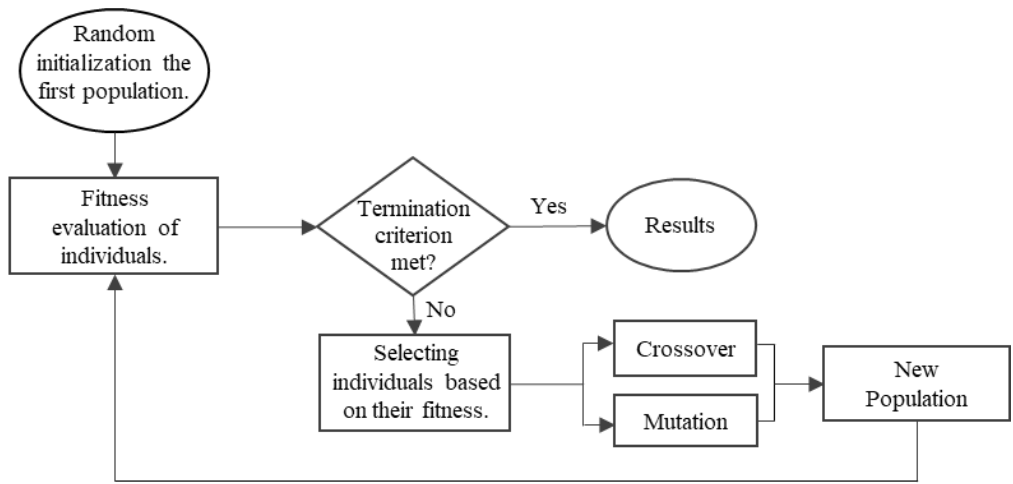

**Figure 1: The flowchart of Genetic algorithm.**

Several studies have evaluated the performance of GAs alongside other optimization techniques such as Particle Swarm Optimization (PSO), Simulated Annealing (SA), and Hooke–Jeeves (HJ) in building optimization problems, consistently showing that GAs can efficiently identify optimal solutions with relatively low computational effort, while other algorithms are more prone to getting trapped in local optima [21], [25], [30]. This makes GAs particularly well-suited for scenarios with moderate computational constraints, where a balance between accuracy and efficiency is essential. However, when computational resources are severely limited, even GAs may struggle to deliver effective results due to risks such as premature convergence and entrapment in local optima [6]. The performance of GA in such cases depends heavily on factors including parameter tuning [4] [25], the fitness landscape [6] [26], and the complexity of optimization variables—suggesting that while GA offers a practical solution under moderate constraints, it may not be ideal under extreme limitations.

On the other hand, when ample computational resources are available, hybrid algorithms often emerge as the most robust option. By combining the strengths of multiple optimization methods, they tend to offer superior accuracy and convergence reliability. For instance, in their evaluation of nine optimization algorithms, Wetter and Wright [21] found that hybrid approaches like PSO-HJ delivered the lowest energy consumption. Nevertheless, GAs remain a strong contender under tighter evaluation budgets, particularly when some trade-off between accuracy and computational cost is acceptable. These observations are echoed across several other studies [22], [23], [24].

This paper explores the effectiveness of GA in early-stage building design optimization by comparing their performance to that of simpler baseline methods like Random Search (RS) and Grid Search (GS) to assess GA's viability under strict computational limitations, through a case study. In some instances, as shown in [31], a simple random sampling has been found to perform comparably to more sophisticated approaches.

RS involves generating random samples within the design space and is particularly useful in high-dimensional problems where variable relationships are complex and not well understood [32]. By sampling randomly, it avoids biases inherent in systematic search methods. While computationally simple, RS can be effective in vast or poorly understood search spaces. In this study, RS serves as a baseline for evaluating GA's performance and assessing the effectiveness of its evolutionary heuristic within the defined case study. GS on the other hand, systematically explores a predefined set of hyperparameters by evaluating every possible combination within the specified grid [33], [34]. This exhaustive approach ensures that all potential configurations are considered, providing a comprehensive understanding of the search space. However, it can be computationally expensive, especially as the number of parameters increases. Despite its computational intensity, GS has the advantage of spanning the solution space evenly. The implementations of all three algorithms are provided in the test case setup section. Figure 2 represents a schematic of search strategies in GA, RS, and GS, respectively labeled from A to C.

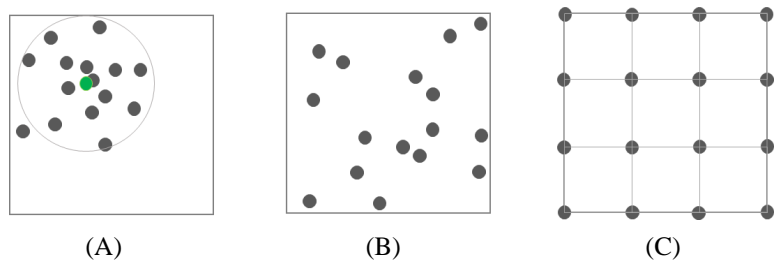

(A)  (B)  (C)

*Figure 2: Schematic illustration of (A): GA, (B): RS, and (C): GS search*

For this evaluation, a specific test case was developed, and an extended run of the GA was conducted until convergence without a time limit to approximate the optimal solution, defined as the minimum annual energy use. This extended run served as the baseline for evaluating the test results. Further details of the test case are provided in Section 3.

# 3 Test Case Setup

The test case represents a typical office building, based on an existing structure located at [*location to be added after review*], with its key characteristics summarized in Table 1. A parametric model is developed using Rhino and Grasshopper, while annual operational energy performance—including heating, cooling, and lighting loads—is simulated using EnergyPlus. Thermal comfort parameters are defined based on ASHRAE Standard 55, and internal load assumptions (e.g., occupancy, electrical equipment, and lighting) are based on the EnergyPlus schedule library for standard office usage. The total annual energy consumption required to maintain thermal comfort and meet internal loads is calculated in kilowatt-hours. To ensure consistent climatic conditions, TMY3 weather data for College Station, Texas—a representative hot and humid location in the United States located in climatic zone 2A—is applied to the model. Figure 3 illustrates the test case model and shows how the building's shape is parameterized using a 4-dimensional vector.

*Table 1: The physical descriptions of the building model.*

| | |
|---|---|
| **Building program** | Office: OpenOffice |
| **Location** | College Station, TX: Climatic zone 2A |
| **Number of floors** | 7 |
| **Area per floor** | 990 m$^2$ |
| **WWR** | 30% |
| **Width to length ratio** | 0.5 |
| **Orientation** | 0º (Y- axis is in North direction) |
| **Opaque construction** | 100mm brick, 200mm heavyweight concrete, 50mm insulation board (Conductivity = 0.03 W/m·K), Wall air space resistance, 19mm gypsum board |
| **Glazing construction** | Clear 3mm, Air 13mm, Clear 3mm |
| **Energy system** | HVAC with Ideal Air Loads (Pre-defined by EnergyPlus) |
| **Operation Schedule** | Academic Calendar - office building |

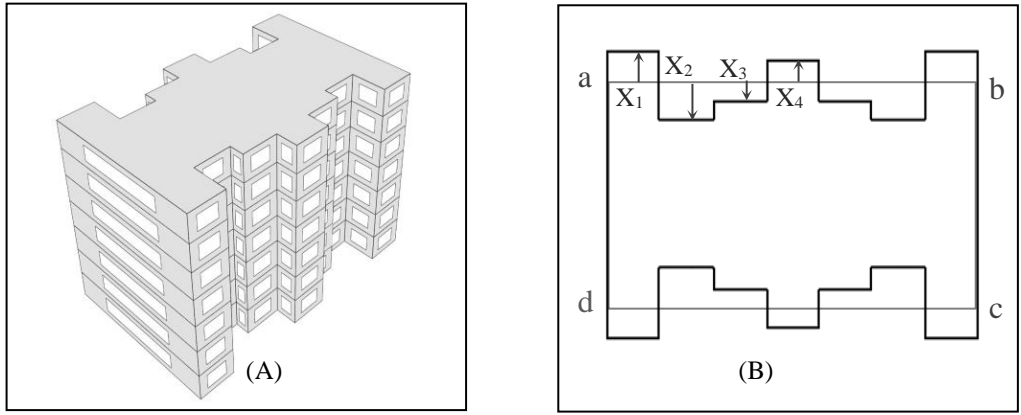

*Figure 3: (A) Building 3-D model. (B) Building shape can be quantified by a 4-dimensional vector.*

It is important to emphasize that the primary objective of this study is to evaluate the effectiveness of GA, GS, and RS methods for early-stage building design optimization, specifically in light of the search space characteristics and imposed computational constraints. To maintain simplicity and focus, the building design is limited to variations in form, while keeping other design variables constant. These include window-to-wall ratio (WWR), building orientation, and envelope thermal properties, which do not directly influence the shape-based search space under investigation. This approach allows for a more targeted and fair assessment of the algorithms' performance within a controlled optimization scenario.

Building form variations are constrained to a specific parametric structure that still generates a sufficiently large and meaningful search space. These constraints preserve a semi-rectangular geometry and exclude complex morphologies such as cross-shaped layouts or irregular forms, ensuring consistency in the architectural layout while maintaining the same floor area as the reference rectangle a-b-c-d illustrated in Figure 3B. A sample of the building form variations explored in the search space is shown in Figure 4**Error! Reference source not found.**. Each variant is defined by a shape vector $X = (x_1, x_2, x_3, x_4)$, with feasible values for each component governed by the parametric expressions described below.

$$X: (x_1, x_2, ..., x_n)$$
$$x_1 + x_2 + \ldots + x_n = 0$$
$$-11.5 \text{ ft} \leqslant x_1, \cdots, x_n \leqslant +11.5 \text{ ft}$$

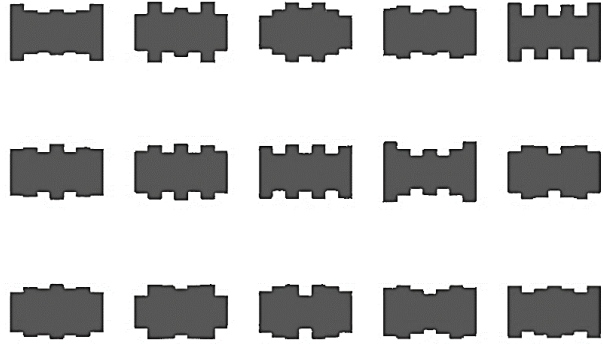

*Figure 4: A sample of variations of the building form in the search space.*

Given the stochastic nature of the optimization algorithms employed, the experiment was repeated ten times to ensure reliable comparisons. Boxplots were used to visualize and compare the results. The near-optimal energy usage, corresponding to the most efficient building designs, was estimated through an extended GA run that continued until convergence without a time limit. This run, which required approximately 75 hours (equivalent to 2,250 simulations), utilized 30 generations, an initial population of 100, and iteration populations of 50. All simulations were performed on a system equipped with an Intel® Core™ i7-1260P CPU (2.10 GHz, 12 cores, 16 logical processors) and 48 GB of RAM.

In the process of optimizing building design, X represents a vector of n shape design decision variables, each with a uniform probability distribution. The components of X are constrained within an n-dimensional feasible region, denoted as S. The primary objective of the optimization algorithm is to identify a value of X within the search space S, such that X∈S, that minimizes the objective function F(X). This function F(X) corresponds to the annual energy consumption of the building and includes three key energy-consuming components: heating, cooling, and lighting/fans, defined as:

$$F(X) = \frac{[Q_h(X) + Q_c(X) + E_{l,f}(X)]}{3.6 \times 10^6}$$

Where:

$Q_h(X)$ and $Q_c(X)$ are the zones' annual heating and cooling consumptions (Joules), respectively, and $E_{(l,f)}(X)$ is the zones' lighting and fans electricity consumption. The denominator is a conversion factor that converts energy from Joules to kilowatt-hours (kWh). The optimal building design, denoted as (X*,Y*) is identified under the constraints of limited computational resources, where:

$$X^* = \arg \min F(X) \quad , X \in S$$

$$Y^* = F(X^*) = \min F(X) \quad , X \in S$$

## 3.1 Search Algorithms Implementations

The GA optimization process begins with a random initialization of the first generation and utilizes crossover and mutation operators to efficiently traverse the search space. pseudocode outlining the GA implementation used in this test case is provided in Table 2.

*Table 2: GA pseudocode implementation for the test case.*

| Genetic Algorithm () | |
|---|---|
| **Step 0:** | **Input:** search space S, iteration index i, initial population: init_pop = 100, iterations population = 50, number of generations: num_gen= 5, number of elitisms: num_elit, number of crossovers: num_cros. |
| | **Output:** $(X^*, Y^*)$ |
| **Step 1:** | **Initialization** <br> for i=1 to init_pop: <br>     $X_i$: $\langle x_1,...,x_n \rangle \in S$ <br>     $f(X_i)$ from EnergyPlus <br>     Save $(X_i, f(X_i))$ in the population Pop; |
| **Step 2:** | **Iteration** <br> loop until the terminal condition: <br>   For i in (1, num_gen): <br>     Select the best (num_elit) samples in Pop and save them in Pop1; |
| **Step 3:** | **Crossover** <br>   Num_cros=(a- num_elit)/2; <br>   For j in (1, number of crossover): <br>     Randomly select two solutions $x_a$ and $x_b$ from Pop; <br>     Generate Xc and $X_d$ by one-point crossover to $X_a$ and $X_b$; <br>     Save $X_c$ and $X_d$ to Pop2; |
| **Step 4:** | **Mutation** <br>   For j=1 to num_cros: <br>     Select a sample $X_i$ from Pop2; <br>     Mutate each bit of Xi under the rate y and generate a new sample Xj2; <br>     If Xj2 is unfeasible <br>       Update Xj2 with a feasible solution by repairing Xj2 |
| **Step 5:** | **Evaluation** <br>   Update Pop = Pop1 + Pop2. <br> return the best solution X in Pop. |

We utilize a uniform and global RS algorithm to optimize building design. Uniform sampling ensures that all solutions are valid and uniformly distributed across the feasible search space. Once a new solution is calculated, it is evaluated, and if it outperforms the previous solution, its value is recognized as the best option; otherwise, the previous solution remains unchanged. The sampling process continues until the termination criterion, 350 iterations, is met. The RS algorithm is developed based on the pseudocode outlined in Table 3.

*Table 3: RS pseudocode implementation for the test case.*

| | **Random Search Algorithm ()** |
|---|---|
| Step 0: | **Input:** search space S, stopping criterion: i <350. |
| | **Output:** (X*, Y*) |
| Step 1: | **Initialization** |
| | $X_0$: $\langle x_1,...,x_n \rangle \in S$ |
| | $f(X_0)$ from EnergyPlus |
| Step 2: | **Iteration** |
| | Generate a random sample $X_i$: $\langle x_1,...,x_n \rangle \in S$ |
| | Objective Function: $f(X_i)$ from EnergyPlus |
| Step 3: | **Evaluation** |
| | If $f(X_0) < f(X_i)$ |
| |     Set $(X^*, Y^*) = (X_0, f(X_0))$ |
| | Else |
| |     Set $(X^*, Y^*) = (X_i, f(X_i))$ |
| | If the stopping criterion is met, stop. |
| | Otherwise increment i and return to Step 1. |

GS uses an exhaustive serach strategy using a pre-defined matrix of variables. However, the required grid for the search surpassed the computational limits established for the study, unless an impractically wide grid was employed. Recognizing these constraints, an alternative strategy was adopted: the selection of solutions from a pre-defined grid at a meaningful distance of 1.6 ft. This random selection process allowed for a more manageable computational load while maintaining the integrity of the exploration process. In response to these considerations, the GS was implemented based on the pseudocode outlined in Table 4.

*Table 4: GS pseudocode implementation for the test case.*

| | **Grid Search Algorithm ()** |
|---|---|
| Step 0: | **Input:** search space S, pre-defined grid of variables, stopping criterion: i <350. |
| | **Output:** (X*, Y*) |
| Step 1: | **Initialization** |
| | $x_1 \in \{k_1, k_1+d_1, k_1+2d_1, ..., k_1+m_1d_1\}$ |
| | $x_2 \in \{k_2, k_2+d_2, k_2+2d_2, ..., k_2+m_2d_2\}$ |
| | $\vdots$ |
| | $x_n \in \{k_n, k_n+d_n, k_n+2d_n, ..., k_n+m_nd_n\}$ |
| | $X_i$: $\langle x_1,...,x_n \rangle \in S$ |
| Step 2: | **Iteration** |
| | Generate a random sample $X'_i \in X_i$ |
| | Calculate $f(X'_i)$ from EnergyPlus |
| Step 3: | **Evaluation** |
| | If $f(X_i) < f(X_{i+1})$ |
| |     Set $(X^*, Y^*) = (X_i, f(X_i))$ |
| | Else |
| |     Set $(X^*, Y^*) = (X_{i+1}, f(X_{i+1}))$ |

### 3.2 Performance Measures

For the comparison of GA, GS, and RS, we adopt three evaluation measures as proposed by Beiranvand et al. [3]: success rate for evaluating reliability and robustness, mean absolute error for assessing accuracy, and lastly the number of simulations representing computational effort for efficiency, with a detailed discussion of these measures provided in the following sections.

(1) **Success Rate:** This measure quantifies the percentage of successful simulations where the optimization algorithm achieves an energy usage within ±0.5% of the estimated minimum energy use. The estimated minimum is determined by running GA without computational constraints until convergence. This margin serves as the success criterion, and the success rate is calculated by dividing the number of near-optimal solutions by the total number of simulations conducted. While this measure indicates how frequently the algorithm approaches the optimal solution within a limited number of simulations, it does not account for the long-term success of the optimization, particularly when extended beyond the initial search.

(2) **Mean Absolute Percentage Error (MAPE):** MAPE is a widely used metric for evaluating the accuracy of prediction models. It provides a relative measure of error by calculating the percentage difference between predicted and actual values. In this context, MAPE compares the optimization results to the near-optimal energy usage derived from the long run of GA, which is considered the benchmark for minimum energy use. The calculation is based on the following equation:

$$MAPE = \frac{100}{n} \sum_{i=1}^{n} \frac{(Y_i - \hat{Y}_i)}{Y_i} \qquad (1)$$

(3) **Computational Effort:** This measure evaluates the computation required to reach a near-optimal solution. To assess the algorithms' efficiency in minimizing computation time, an early threshold (k) is set, specifying how many times the algorithm must meet the success criteria before the threshold is considered met. Setting k to 5 ensures that an optimal solution is captured. This metric also helps identify potential disadvantages of GA if it deviates from the optimal solution. Computational effort is defined as the number of simulations required to meet the threshold for k successful instances.

## 4   Test Case Fitness Function Landscape

To better understand the optimization challenges involved, we examine the fitness function landscape of the test case described in Section 3. When evaluating diverse building shape configurations, the search space reveals a high degree of complexity and noise, which significantly influences the behavior of optimization algorithms. Figure 5 presents a series of plots illustrating the fitness function landscape, each depicting the relationship between two of the four design variables ($X_1$, $X_2$, $X_3$, and $X_4$) and the corresponding annual energy consumption. These plots reveal the complex and irregular nature of the search space, characterized by multiple local optima, which can impede the optimization algorithm's ability to consistently identify the global minimum. This complexity poses a significant challenge: the presence of noise and a scattered distribution of near-optimal solutions increases the likelihood of algorithms converging prematurely to suboptimal results. The observed landscape underscores the importance of selecting or designing optimization strategies capable of navigating such environments effectively—especially under tight computational constraints.

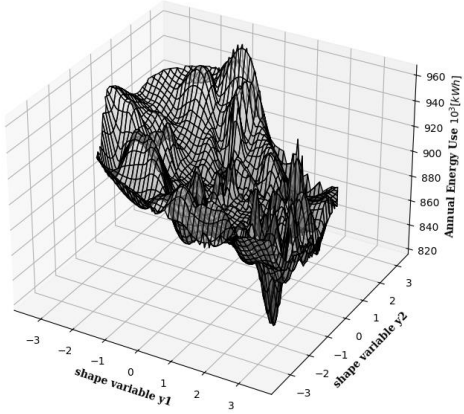
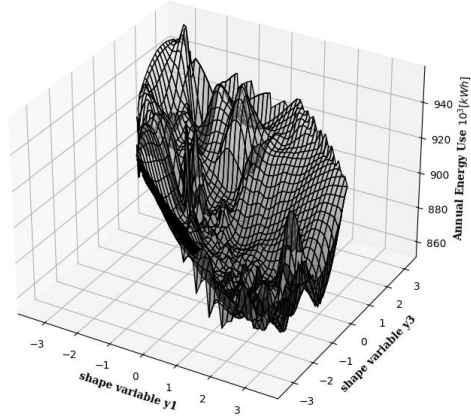
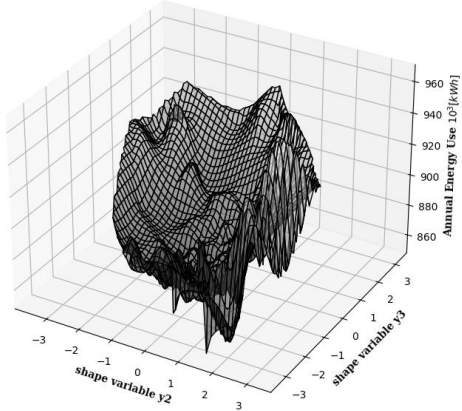
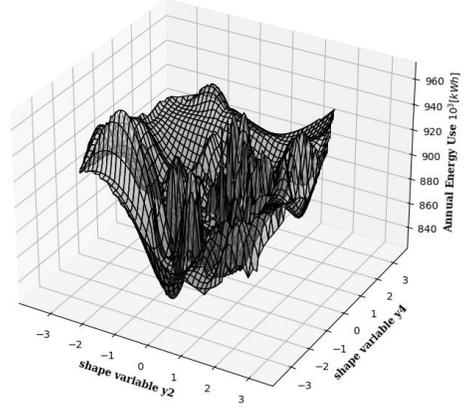
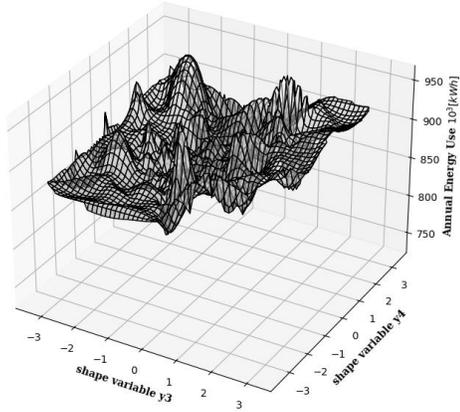
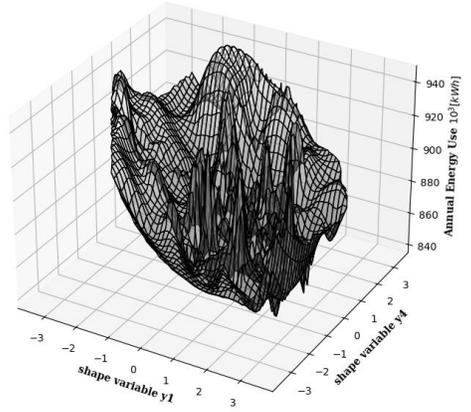

**Figure 5: Illustrations of fitness landscapes in 3D with fitness value (Energy Usage) on Z-axis.**

## 5 Discussion and Results

This section presents the results and discussion of the optimization algorithms in terms of their reliability and robustness (success rate), accuracy (mean absolute percentage error), and efficiency (computational effort), providing insights into their performance across these key metrics.

### 5.1 Reliability and Robustness: Success Rate

To begin, we assess the performance of the three examined methods by comparing their success rates, which measure how often the search algorithm approaches an optimal solution within a restricted number of simulations. This metric ranges from 0 to 100, where 0 indicates no success, 100 represents a fully successful search where all trials converge to an optimal or near-optimal solution, and any non-zero value indicates successful performance. Given the relatively small number of near-optimal solutions in the search space, we expect this value to be low, but a higher value suggests more frequent convergence toward the optimal solution. Boxplots illustrating the repetition of this experiment are shown in Figure 6. On average, the success rates of the three algorithms within the set constraint are similar, with GA often being the best performer in individual repetitions. However, a closer examination of the medians (depicted by the white line) shows that when considering all repetitions, RS outperforms GA overall. In some repetitions, GA failed to find any near-optimal solutions, as shown by the boxplots containing zero. In contrast, the boxplot for RS does not contain zero and starts at 0.94, indicating that RS consistently found near-optimal solutions across all repetitions, resulting in a higher overall success rate.

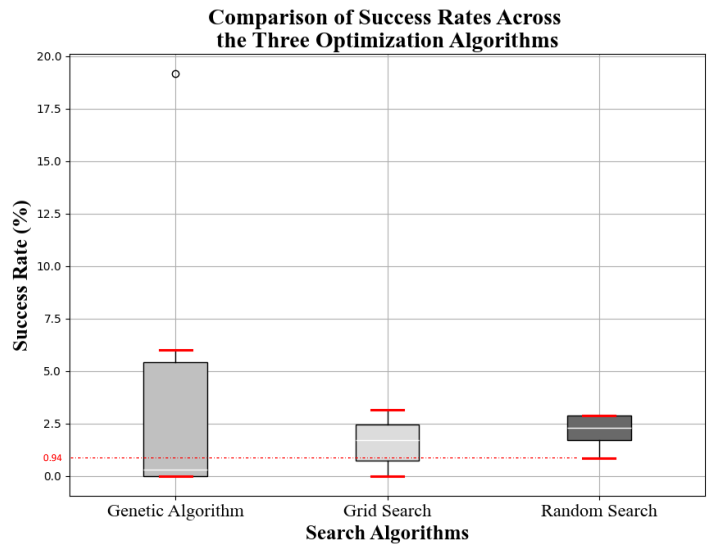

*Figure 6: Boxplots of success rates for GA, GS, and RS optimization methods across all test case repetitions.*

This discovery is significant, emphasizing that when confronted with problems of substantial size, complexity, and noise in the search space, although GA excels over extended periods, RS may yield superior results when the number of simulations is limited. It is noteworthy that, in this specific experiment, RS directs its search efforts globally across the entire search space. In contrast, GA predominantly conducts its global search during the initial algorithm initialization and then allocates a significant portion of its search effort to the local state. GS, lacking heuristic capabilities to efficiently span its search effort across the entire search space, positions itself between RS and GA in terms of success rate. Additionally, RS exhibits a significantly smaller standard deviation for success rates, as shown in the boxplots, suggesting greater consistency in achieving similar results.

## 5.2 Accuracy: Mean Absolute Percentage Error

Subsequently, we will assess the proficiency of the three methods in terms of accuracy in predicting the minimum energy use reported by their mean absolute percentage errors. Figure 7 offers a comprehensive overview of the optimization outcomes across all experiment repetitions. Upon closer examination, it becomes apparent that GA exhibits superior performance compared to both GS and RS, although the medians of the results are relatively similar. This observation implies that when GA is not misled, it tends to deliver optimal results. However, this occurrence may be infrequent, as evidenced by multiple repetitions where GA failed to identify an optimum solution. Conversely, RS consistently identifies optimum solutions, albeit not surpassing the excellence of GA results. Nonetheless, RS proves to be valuable in the initial phases of the design process, where obtaining a satisfactory solution is pivotal.

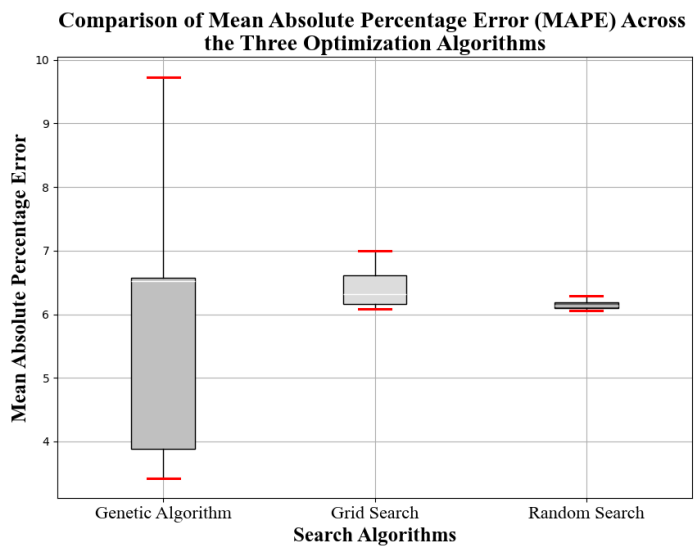

*Figure 7: Boxplots showing the Mean Absolute Percentage Error (MAPE) for the GA, GS, and RS optimization methods across all test case repetitions.*

## 5.3 Efficiency: Computational Effort

Finally, we evaluate the three methods based on the computational effort required, measured in terms of the number of simulations needed to reach a near-optimal solution. To address a known limitation of GA—its occasional tendency to deviate from the optimal path—we introduce a success threshold ($k$), set to 5, ensuring that an optimal solution is reliably captured multiple times. Figure 8 presents the boxplots illustrating the simulation effort required by each algorithm to achieve $k$ successful outcomes. On average, our experiment indicates that the three algorithms required similar computational resources. However, the medians for GA and GS are notably lower than RS, suggesting they can discover a near-optimal solution more quickly in certain repetitions. GA may perform better when it is heading in the right direction, enabling it to reach a result faster. However, if misled, GA can fail to converge to an optimal solution, which introduces variability in its performance. On the other hand, RS, despite its stochastic nature, consistently provided near-optimal solutions in all repetitions, making it a more reliable choice under the set computational constraints. This highlights that, although GA's heuristic approach can sometimes yield faster results when it is in the right direction, it doesn't suffice when computational resources are limited, making RS potentially a better option in such cases.

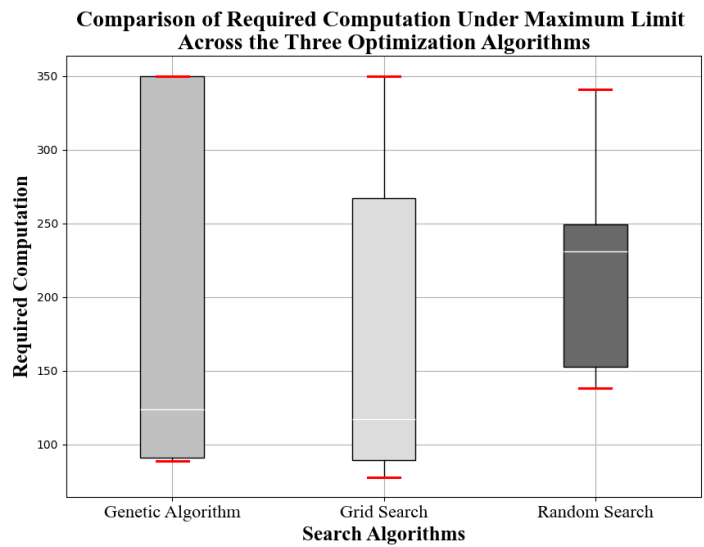

*Figure 8: Boxplots of required computation for GA, GS, and RS optimization methods across all test case repetitions.*

## 6   Conclusion and Future Work

This study assessed the effectiveness of Genetic Algorithms (GAs) for early-stage building design optimization, specifically evaluating their ability to identify near-optimal solutions under constrained computational budgets. Using a single case study, we compared the performance of GA with two baseline methods—Random Search (RS) and Grid Search (GS)—yielding several key insights. In terms of success rate, RS demonstrated comparable averages to GA and GS but with significantly lower standard deviation, indicating greater consistency across repeated runs. When evaluating accuracy through mean absolute percentage error (MAPE), GA showed superior precision in predicting minimum energy usage. However, it also exhibited occasional failures to converge within the given computational constraints. Conversely, RS consistently found near-optimal solutions, highlighting its potential reliability in early design stages, although its stochastic nature cannot guarantee optimality in every case. In this experiment, RS maintained a broad search across the design space throughout, whereas GA concentrated its global search effort in the early stages before shifting focus to local refinement. GS, lacking heuristics for strategic exploration, fell between RS and GA in terms of overall effectiveness. These findings suggest that in scenarios with strict computational limitations, simpler methods like RS—or potentially hybrid strategies—may outperform more complex algorithms such as GA. As this study is limited to a single case under specific constraints, future work should expand this research to a broader range of design scenarios and computational settings, and further explore the potential of hybrid or adaptive methods that leverage the strengths of multiple algorithms for improved performance and generalizability.